\documentclass[10pt,twocolumn,letterpaper]{article}
\usepackage[T1]{fontenc}
\usepackage[pagenumbers]{cvpr}

\usepackage{titletoc}
\usepackage{multicol}
\usepackage{amsmath,amssymb}
\usepackage{graphicx}
\usepackage{booktabs,multirow,array}
\usepackage{colortbl,tabularx}
\usepackage{placeins}
\usepackage{microtype}
\usepackage{enumitem}
\usepackage{xspace}
\usepackage{fancyhdr}
\newcommand{\method}{SpatialCast\xspace}

\graphicspath{{figures/}}

\definecolor{cvprblue}{rgb}{0.21,0.49,0.74}
\definecolor{oursblue}{RGB}{232,242,250}
\definecolor{ourslavender}{RGB}{240,240,255}
\usepackage[pagebackref,breaklinks,colorlinks,allcolors=cvprblue]{hyperref}
\def\confName{}
\def\confYear{}

\title{SpatialCast: Reading the 3D World via Cast Shadows}

\author{Enhan Li\textsuperscript{1} \quad Jinghao Zhou\textsuperscript{1} \quad Jiaze Zhang\textsuperscript{1} \quad Hongyang Du\textsuperscript{1,}\thanks{Corresponding author.}\\
\normalfont\normalsize\textsuperscript{1}The University of Hong Kong
}

\fancypagestyle{labpaperfirst}{
  \fancyhf{}
  \fancyhead[L]{%
    \raisebox{6mm}[0pt][0pt]{%
      \includegraphics[width=4.2in]{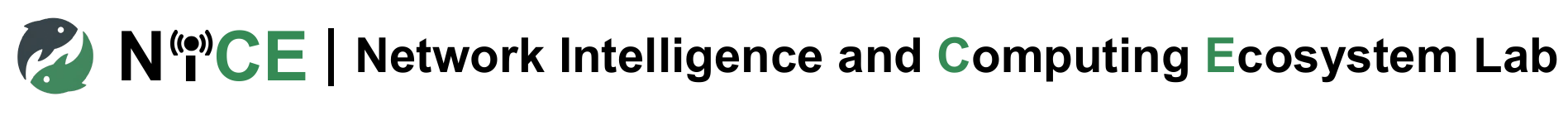}%
    }%
  }
  
}

\begin{document}
\maketitle

% labpage
\thispagestyle{labpaperfirst}

% ===== sec/0_abstract =====
\begin{abstract}
As multimodal large language models (MLLMs) advance toward understanding and
interacting with the physical world, spatial reasoning becomes essential.
Existing spatial benchmarks have made substantial progress in evaluating spatial
relations, memory, and viewpoint reasoning, but  have less systematically assessed how
models connect physical scene conditions with visual observations.
We introduce \method, a benchmark that uses cast shadows to evaluate this
ability.
Cast shadows provide observable evidence of a 3D scene.
Their appearance in images depends on object geometry and pose, illumination,
and viewpoint.
We examine this relationship through three complementary tasks that assess
scene--projection consistency, predict intervention-induced changes, and
infer the physical changes behind visual transitions.
The benchmark contains 21,000 synthetic questions with separate
scene and geometry test partitions, complemented by 200 SpatialCast-Real
questions from 20 photographed scenes.
Evaluations show that open-weight MLLMs, including spatially trained models, remain near chance on the synthetic benchmark, whereas closed-source models perform substantially better but still struggle with shadow-length and light-height judgments.
We also use \method training data for GRPO post-training, improving
open-weight MLLM performance on the benchmark.
Further evaluations show improvements on several external spatial benchmarks,
suggesting transfer of the learned capabilities.
These experiments support \method as a controlled resource for
diagnosing spatial reasoning capabilities and studying their learning
and generalization.
\end{abstract}

% ===== sec/1_intro =====
\section{Introduction}
\label{sec:intro}

\begin{figure*}[t]
    \centering
    \includegraphics[width=\textwidth]{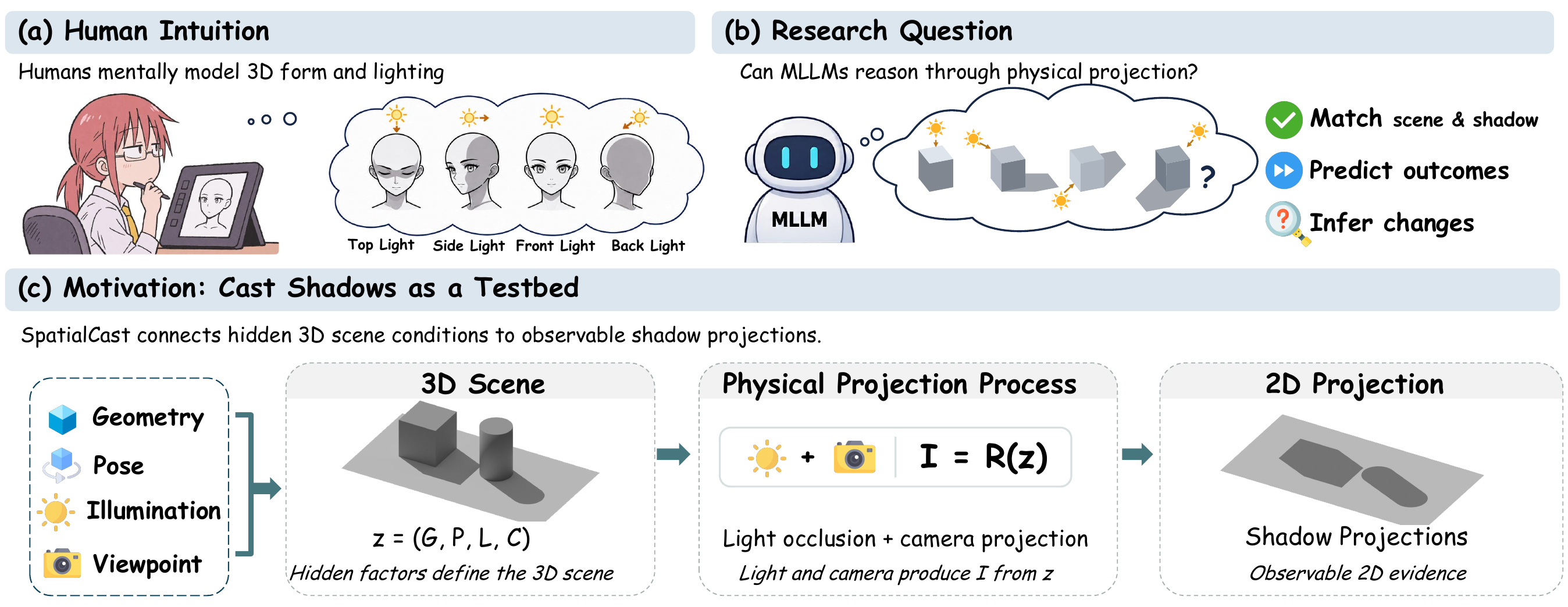}
    \caption{\textbf{Motivation for SpatialCast.}
    (a) Human intuition for relating 3D form and lighting to visual appearance.
    (b) We ask whether MLLMs can match scenes with shadow projections,
    predict intervention outcomes, and infer change parameters from paired
    observations given the changed factor.
    (c) Cast shadows provide physically constrained evidence connecting
    scene geometry, pose, illumination, and viewpoint to observable 2D projections.
    The shadow footprints are schematic; benchmark observations contain
    both objects and their cast shadows.}
    \label{fig:motivation}
\end{figure*}

Multimodal large language models (MLLMs) are increasingly expected to
move beyond interpreting images and videos toward understanding and
interacting with the physical world~\cite{zheng2025multimodalspatialsurvey}.
Such interactions require models to understand spatial structures and
anticipate how visual observations change as an observer moves or a scene is
modified~\cite{spatialvlm2024,liu2025spatialreasoningsurvey}.
Evaluating these abilities is an important step toward MLLMs that
understand and interact with the 3D world.
Moreover, recent perspectives on spatial intelligence have highlighted
new-view prediction as a visual counterpart to next-token prediction
in language modeling~\cite{li2026worldmodelsdiscussion}, underscoring the
importance of understanding how scene conditions give rise to visual
observations.
However, viewpoint is only one of several physical factors
that shape what we see.
Changes in illumination and object pose also alter
visual observations, calling for a broader evaluation of how physical scene
conditions determine their 2D projections.

Existing benchmarks have examined geometric relationships, including relative
positions, distances, and orientations~\cite{spatialvlm2024,spatial4572025}, as well as spatial memory, perspective-taking, and mental simulation
from multiple or limited observations~\cite{yang2025thinking,mindcube2025}.
More recent work extends these evaluations to dynamic settings, examining the effects of object and observer
movements~\cite{sat2025}, projection and viewpoint
transformations~\cite{spatialblock2026}, and spatially constrained image
editing~\cite{gsi2026}.
However, existing benchmarks less systematically test how physical changes
alter visual observations. They also provide limited evaluation of whether
models can infer these changes from before-and-after observations.
Cast shadows offer a natural way to probe this relationship.
As shown in Fig.~\ref{fig:motivation}, the appearance of a cast shadow depends on object geometry,
pose, illumination, and their combined effects on the resulting projection.
Cast shadows therefore provide observable evidence about the underlying
3D scene.
By controlling the physical scene, we can systematically evaluate
whether models understand the resulting projections, anticipate their changes under interventions, and infer the physical changes behind observed transitions.

Building on this insight, we introduce \method, a benchmark that uses
controlled rendering to evaluate how MLLMs reason about the physical
relationship between 3D scenes and their 2D projections.
By varying scene conditions while preserving their underlying physical
constraints, \method provides verifiable questions that probe this
relationship from three complementary directions.
\emph{Projection Compatibility} (T1) tests agreement between scene
conditions and projections through forward prediction and inverse
inference of lighting.
\emph{Projection Propagation} (T2) predicts the
projection after a specified light, object-rotation, or viewpoint change.
\emph{Projection Attribution} (T3) infers change parameters from paired
observations, given the factor that changed.
The benchmark contains 21,000 synthetic questions across
16 shape families and three scene structures, with separate scene and
geometry test partitions.
SpatialCast-Real adds 200 questions from
20 photographed ReNe scenes to assess light-change reasoning in real
images~\cite{toschi2023relight}.

Our contributions are as follows.
\begin{itemize}
    \item \textbf{Cast shadows as a physical readout for spatial reasoning.}
    We leverage cast shadows as observable evidence of the underlying 3D
    scene, whose image appearance is governed by object geometry, pose,
    illumination, and viewpoint.
    We design three complementary tasks to
    evaluate scene--projection compatibility, predict the visual consequences
    of physical interventions, and infer changes from before-and-after
    observations.

    \item \textbf{A controlled benchmark spanning synthetic and real scenes.}
    We use BlenderProc and Blender Cycles to construct controllable 3D
    scenes and generate 21,000 synthetic questions across 16 shape families
    and three scene structures, with separate test partitions for new scene
    instances and held-out geometric parameter ranges.
    We further introduce SpatialCast-Real, comprising 200 questions based
    on photographed lighting changes.

    \item \textbf{Extensive evaluation and generalization.}
    We evaluate SpatialCast on both proprietary and open-weight MLLMs,
    including spatially trained baselines, revealing distinct performance
    profiles across models and reasoning tasks.
    SpatialCast-GRPO training yields model-dependent improvements,
    including a 23.33-percentage-point gain on Projection Propagation
    for the 8B model.
    The trained models also show improvements on several external spatial
    benchmarks, demonstrating the potential for cross-benchmark
    generalization.
\end{itemize}

% ===== sec/2_related =====
\section{Related Work}
\label{sec:related}

\paragraph{Spatial Intelligence in MLLMs.}
SpatialPIN and Thinking with Blueprints incorporate 3D priors or structured
object representations~\cite{spatialpin2024,blueprints2026}, while
related methods use geometric primitives and bounding-box
abstractions~\cite{primitives2026,sandboxvlm2026}.
Other work improves spatial reasoning through metric
supervision~\cite{spatialvlm2024}, geometric surrogate
tasks~\cite{euclid2026,sparkle2025}, and verifiable objectives based on
self-supervised spatial tasks or cross-view
matching~\cite{spatialssrl2026,reasonmatch2026}.
SpatialLadder and Spatial-MLLM further explore progressive training and
geometry-aware visual features~\cite{spatialladder2025,spatialmllm}.
Several of these methods also provide spatially trained baselines for
examining whether learned capabilities transfer to different spatial
reasoning tasks.

\paragraph{Benchmarks for Spatial Intelligence.}
Existing benchmarks differ in their visual observations and reasoning
targets.
Spatial457 and OmniSpatial evaluate diverse spatial
abilities~\cite{spatial4572025,omnispatial2026}, while VSI-Bench and MindCube
examine spatial reasoning from videos or limited
views~\cite{yang2025thinking,mindcube2025}.
Multi-SpatialMLLM and MMSI-Bench further investigate spatial understanding
across multiple frames or images~\cite{multispatial2026,mmsibench2026}.
Controlled settings enable more targeted evaluations, including state
reconstruction and transformations in cube-based
benchmarks~\cite{cubebench_anand2025,cubebench_gao2025} and procedural scene
complexity in InfiniBench~\cite{infinibench2026}.
More directly related to scene changes, SAT examines how object and
observer movements affect spatial relationships~\cite{sat2025}, while
SpatialBlock evaluates 3D-to-2D projection, viewpoint transformation, and
structural combination through synthetic block-stacking
tasks~\cite{spatialblock2026}.
GSI-Bench assesses whether generated image edits satisfy specified spatial
constraints~\cite{gsi2026}, whereas CamChoreo identifies camera-motion types
and temporal intervals in videos~\cite{du2026camchoreo}.
These benchmarks primarily examine spatial relationships, transformations,
or visual outputs, while the physical relationship between scene changes
and their observable consequences offers a complementary perspective for
evaluating spatial reasoning.

\paragraph{Shadows and Geometric Reasoning.}
Cast shadows provide geometric evidence about the underlying 3D scene, as
their formation depends on object geometry, pose, and illumination.
Prior work exploits shadows to reconstruct plausible hidden object shapes
while jointly estimating pose and lighting, showing that multiple scene
explanations may be consistent with the same
observation~\cite{shadowreconstruct2023}.
Differentiable shadow mapping further enables gradient-based optimization
of physical scene parameters~\cite{diffshadow2023}.
Other studies investigate shadow detection from a vision-language
perspective~\cite{shadowvl2026} or optimize object pose and illumination for
shadow-based visual compositions~\cite{shadowdraw2026}.
ReNe provides real photographs with calibrated light and camera poses for
relighting and novel-view synthesis~\cite{toschi2023relight}.
The ambiguity of shadow-based reconstruction also highlights the importance
of specifying known scene conditions and candidate hypotheses in inverse
reasoning tasks.

% ===== sec/3_dataset =====
\section{SpatialCast Benchmark}
\label{sec:dataset}

\begin{figure*}[t]
    \centering
    \includegraphics[width=\textwidth]{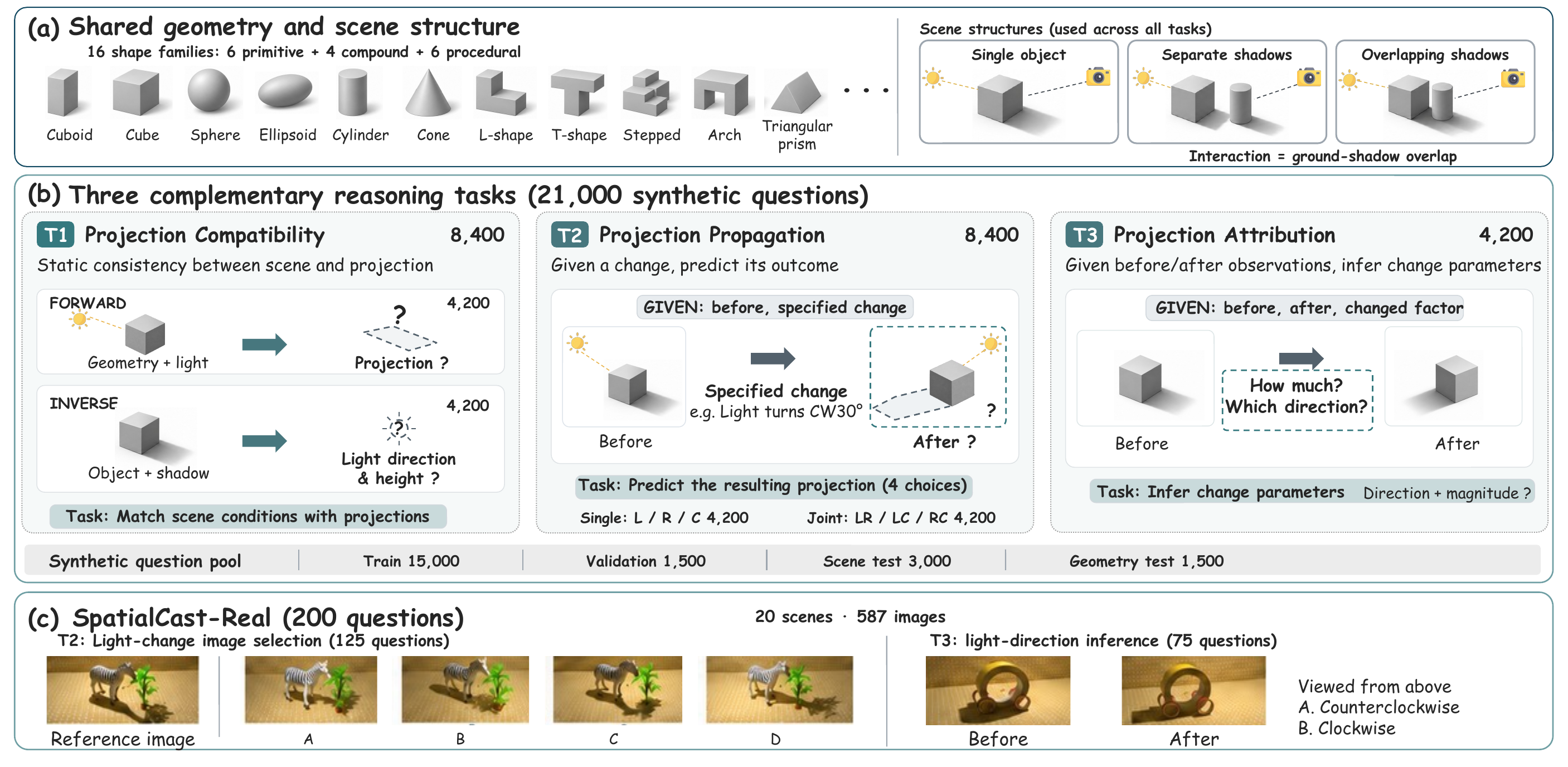}
    \caption{\textbf{SpatialCast task and dataset overview.}
    Panels (a,b) are schematic illustrations.
    (a) The tasks share 16 shape families and three scene structures.
    (b) T1 combines forward prediction and inverse
    inference of lighting.
T2 predicts the outcome of a specified intervention.
T3 infers change parameters from before/after observations and a known
    changed factor.
The synthetic question pool contains 21,000 questions,
    including 4,500 test questions.
(c) SpatialCast-Real contains
    200 questions across 20 photographed scenes, illustrated by a T2
    reference with four candidate outcomes and a T3 before/after pair.
Counts refer
    to the complete pool before training-data screening.
Experiments use
    the full T1/T2/T3 test scope shown here.}
    \label{fig:dataset}
\end{figure*}

\method tests three inference directions through a shared projection
map.
Figure~\ref{fig:dataset} summarizes how the synthetic benchmark is
constructed and partitioned, alongside examples of real-image questions.

\subsection{Scene Representation}
\label{sec:scene}

A scene state $z=(G,P,L,C)$ specifies object geometry $G$, object poses
and positions $P$, illumination $L$, and camera configuration $C$ (Fig.~\ref{fig:motivation}).
A renderer $\mathcal{R}$ implements the physical projection process,
producing the object--shadow observation
\begin{equation}
    I=\mathcal{R}(G,P,L,C).
    \label{eq:render}
\end{equation}
Scenes contain a planar shadow receiver and one or two objects.
An
intervention $a$ updates specified components of the state, producing
$z'=T_a(z)$ and observation $I'=\mathcal{R}(z')$, where $T_a$ is the
state transformation specified by $a$.
Light changes and object rotations affect shadow formation.
Camera
changes alter the view of the resulting scene.

\subsection{Geometry and Scene Construction}
\label{sec:construction}

The geometry pool contains six primitive, four compound, and six procedural
shape families, totaling 16 (Fig.~\ref{fig:dataset}(a)).
Their silhouettes and exposed surfaces
produce varied responses to light and pose changes.
Supplementary
Sec.~\ref{app:a1} lists all families with example geometry instances.

The three scene structures used in each task contain a single object, two
objects with separate ground shadows, or two objects with overlapping ground
shadows.
Overlap introduces a compositional observation in which the
visible shadow boundary can depend on both objects.
The same structure
labels apply across tasks, supporting task-level and structure-level
analysis without changing the task definitions.
Scenes are rendered with BlenderProc and Blender
Cycles~\cite{blenderproc2023}.

\subsection{Three Complementary Tasks}
\label{sec:tasks}

The three tasks in Fig.~\ref{fig:dataset}(b) share the scene representation
and physical projection model of Sec.~\ref{sec:scene}.
T1 matches scene conditions to projections within a static scene.
When the scene changes, T2 predicts the resulting projection given the
intervention, while T3 uses the before and after observations to infer
how the specified factor changed.
Table~\ref{tab:task_overview} summarizes the inputs, prediction targets,
and variants of each task.

\begin{table}[!ht]
\caption{\textbf{Overview of the three SpatialCast reasoning tasks.}
L, R, and C denote light, object rotation, and camera viewpoint,
respectively. Paired letters denote joint interventions.}
\label{tab:task_overview}
\centering
\footnotesize
\setlength{\tabcolsep}{2.5pt}
\renewcommand{\arraystretch}{1.08}
\begin{tabularx}{\columnwidth}{@{}>{\raggedright\arraybackslash}p{.20\columnwidth} >{\raggedright\arraybackslash}X >{\raggedright\arraybackslash}X >{\raggedright\arraybackslash}p{.19\columnwidth}@{}}
\toprule
\textbf{Task} & \textbf{Given} & \textbf{Select} & \textbf{Variants} \\
\midrule
\multirow[t]{2}{=}{\textbf{T1}\newline Compatibility}
  & Geometry and light & Compatible projection & Forward \\
\cmidrule(l){2-4}
  & Observed shadow & Compatible light setting & Inverse \\
\midrule[0.35pt]
\textbf{T2}\newline Propagation
  & Initial view $I$ and intervention $a$ & Resulting view $I'$
  & L, R, C\newline LR, LC, RC \\
\midrule[0.35pt]
\textbf{T3}\newline Attribution
  & Views $I, I'$ and the changed factor & Parameters of the change & L, R, C \\
\bottomrule
\end{tabularx}

\end{table}

\begin{figure*}[t]
    \centering
    \includegraphics[width=0.96\textwidth]{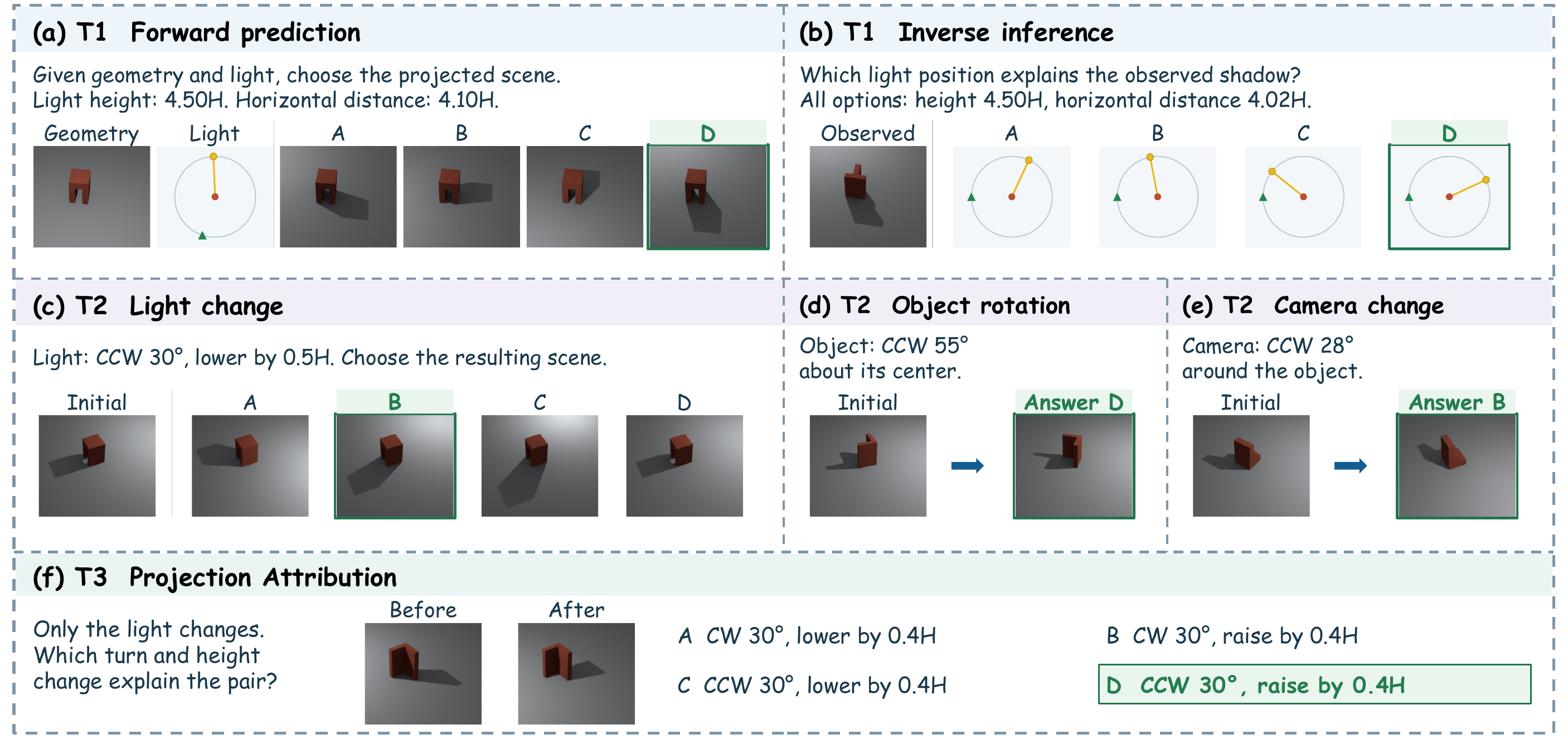}
    \caption{\textbf{Representative SpatialCast questions.}
    (a,b) T1 asks for a projected scene or a light configuration.
    (c--e) T2 specifies a light, object-rotation, or camera change.
    Panel (c) shows the initial observation and all A--D candidates.
Panels
    (d,e) show the initial observation and the correct candidate.
    (f) T3 gives the changed factor and asks for its change parameters.
    Green highlights mark ground-truth answers.
In the top-view diagrams,
    yellow circles denote the light, red circles the object, and green
    triangles the camera.
$H$ denotes object height.
CW and CCW refer to
    clockwise and counterclockwise turns viewed from above.
Scene views share a fixed display
    crop within each question.}
    \label{fig:question_examples}
\end{figure*}

\paragraph{T1: Projection Compatibility.}
T1 tests static compatibility by asking the model to select the physically
consistent candidate from four alternatives.
In the forward variant (Fig.~\ref{fig:question_examples}(a)), the model
selects the projection consistent with the given geometry and light,
using shadow direction, length, or contour.
In the inverse variant (Fig.~\ref{fig:question_examples}(b)), it selects
the light setting (azimuth or height) consistent with the observed shadow
under fixed geometry and viewing conditions.

\paragraph{T2: Projection Propagation.}
T2 asks how an initial object--shadow observation changes under a
specified intervention.
Given the initial image $I$ and intervention $a$, the model selects the
resulting view $I'=\mathcal{R}(T_a(z))$ from four candidate images.
Single-factor interventions change the light (L), rotate an object (R),
or change the camera viewpoint (C). Joint interventions combine two factors (LR, LC, or RC).
For scenes with two objects, the question identifies the object
affected by a rotation.
In Fig.~\ref{fig:question_examples}(c), the light turns $30^\circ$
counterclockwise and drops by half the object's height. Figure~\ref{fig:question_examples}(d,e) shows object rotation and
camera viewpoint changes, respectively.
Supplementary Sec.~\ref{app:question_examples} gives complete sets for every T2 type.

\paragraph{T3: Projection Attribution.}
T3 asks which change parameters explain a transition between
object--shadow observations, given the factor that changed.
In Fig.~\ref{fig:question_examples}(f), only the light changes. From the
before-and-after observations, the model infers whether the light turned
clockwise or counterclockwise and whether it rose or fell.
Object-rotation and camera-viewpoint questions ask about the direction and
magnitude of rotation.
Providing the changed factor defines the hypothesis space for inferring the corresponding change 
parameters from the observed transition.

\paragraph{Question and candidate construction.}
Each synthetic question has four options. Scene metadata identifies the
correct option, and models see only the images, prompt, and options.
T1 fixes the objects and camera. For forward questions, we render the
correct option under the light setting specified in the prompt and the
three distractors under three alternative light settings.
Inverse questions ask which of four light settings produced the observed
image, with the options varying either in azimuth at a fixed height or
in height at a fixed azimuth.
T2 renders the specified intervention and three alternatives from the
same initial scene. Distractors omit a factor, reverse a turn, or
change its magnitude.
T3 pairs each turn direction with a rise or fall for the light, or with
two rotation angles for the object or camera. The four combinations form
the text options, and one is sampled to generate the after image.

\paragraph{Quality control.}
During generation, we reject scenes if too little of an object or its
ground shadow is visible, or if a ground shadow enters the frame margin.
For two-object scenes, we measure ground-shadow overlap to verify the
specified scene structure.
For T3, we check that the four text options cover all combinations of
two binary attributes, so each attribute value occurs twice.
We render the scene for each of the four physical
hypotheses, including for inverse T1 and T3 questions, whose answer
options are lighting diagrams and text descriptions, respectively.
For each of the six pairs of scene renders, we count pixels in the
object or shadow region whose color difference meets a fixed threshold.
For 97.87\% of synthetic questions, even the closest pair differs at
150 or more such pixels (Supplementary Sec.~\ref{app:quality}).

\subsection{Dataset Composition and Partitions}
\label{sec:splits}

The synthetic benchmark contains 21,000 questions.
Of the 8,400 T1
questions, 4,200 use forward prediction and 4,200 use inverse inference.
The 8,400 T2 questions are evenly divided between single-factor and
joint interventions.
The 4,200 T3 questions are approximately balanced
across light changes, object rotations, and camera viewpoint changes.
The dataset is partitioned into 15,000 training, 1,500 validation,
3,000 scene-test, and 1,500 geometry-test questions as summarized in Fig.~\ref{fig:dataset}(b).

\textbf{Scene-test} evaluates generalization to unseen scene instances
independently sampled from the same scene distribution as the training set.
\textbf{Geometry-test}
evaluates generalization to held-out geometric parameter ranges within
the same shape families.
The two test partitions contain 4,500 questions, with 1,800 T1,
1,800 T2, and 900 T3 questions (a 2:2:1 ratio).
Supplementary Secs.~\ref{app:split}--\ref{app:quality} provide detailed counts and
split audits.

\subsection{SpatialCast-Real}
\label{sec:real_data}

To complement the synthetic evaluation, we construct SpatialCast-Real from
ReNe (\emph{ReLight My NeRF})~\cite{toschi2023relight}, a photographic
dataset with calibrated camera and light poses.
Each photograph is
captured with one light on at a time.
We group photographs by scene and camera pose, keeping object geometry,
layout, and viewpoint fixed while varying illumination.
From the calibrated
light poses, we compute the change in lighting between photographs and
use it to label each question.

SpatialCast-Real evaluates the light-change variants of T2 and T3
(Fig.~\ref{fig:dataset}(c)).
For T2, the model receives a reference photograph and the specified
changes in light azimuth, height, and horizontal distance.
It selects
the photograph that matches these changes from four candidates.
For T3, the model compares before-and-after photographs and decides whether
the shortest turn between the two light azimuths is clockwise or
counterclockwise.
A top-view diagram establishes the coordinate convention.
The dataset contains 200 questions across 20 scenes (125 T2 and 75 T3,
ten per scene).
Its 587 public images comprise 585 distinct photographs
and two coordinate diagrams.
A photograph can appear in multiple questions.
We evaluate all 200 questions separately from the synthetic benchmark.
Supplementary Sec.~\ref{app:real_construction} details the construction, and Fig.~\ref{fig:real_question_examples} shows complete
T2 examples in panels~(a,b) and T3 examples in panels~(c,d).

% ===== sec/4_experiments =====
\section{Experiments}
\label{sec:experiments}

We first compare benchmark performance across models and examine learning
across the three reasoning directions.
We then validate the benchmark through controlled input ablations and
human evaluation before examining cross-benchmark transfer.

\subsection{Experimental Setup}
\label{sec:experimental_settings}
\label{sec:protocol}

\paragraph{Compared Models.}
Our four base models are Qwen3-VL-8B Base and Qwen3-VL-4B Base~\cite{qwen3vl2025},
Qwen2.5-VL-3B Base~\cite{qwen25vl2025}, and InternVL3-2B Base
(Instruct checkpoint)~\cite{internvl32025}.
Spatially trained baselines include Spatial-SSRL~\cite{spatialssrl2026},
SpatialBlock~\cite{spatialblock2026}, Spatial-MLLM~\cite{spatialmllm},
and SpatialLadder~\cite{spatialladder2025}.
We evaluate all open-weight models, including our SpatialCast-GRPO
variants, on the 4,500 synthetic test questions and 200 SpatialCast-Real
questions.
We also evaluate GPT-6 Astra, GPT-6.1 Sol, Claude Sonnet 5.5, Gemini 4
Argon, and Qwen3.8-Max (0902) on SpatialCast-Real.
Of these, GPT-6.1 Sol and Gemini 4 Argon also complete the full
synthetic test set.

\paragraph{Evaluation Protocol and Metrics.}
\label{sec:evaluation}
Each model receives the question images, prompt, and answer options.
We use exact-match accuracy, counting unresolved or invalid answers as
incorrect.
We compute accuracy separately for the synthetic test set and
SpatialCast-Real, weighting each question equally within its test set.
The synthetic overall score is micro-averaged accuracy, with T1, T2,
and T3 contributing in a 2:2:1 ratio.
The uniform random-choice baselines are 25\% for the synthetic tasks
and 34.375\% for SpatialCast-Real.
We compute accuracy gains on \method from integer correct counts before
rounding.
Supplementary Sec.~\ref{app:small_model_evaluation} details model-specific inference settings
and answer parsing.

\paragraph{SpatialCast-GRPO Post-training.}
\label{sec:training}
We train each of the four base models from its original weights using
Group Relative Policy Optimization (GRPO)~\cite{deepseekmath2024}.
After screening the training split, we retain 11,186 questions, comprising
1,189 forward T1, 1,998 inverse T1, 6,000 T2, and 1,999 T3 questions.
All four models use the same data order and train for 1,399 updates.
At each update, we sample eight responses per question and reward answer
correctness and format compliance, with a length penalty.
We use each model's final checkpoint for evaluation.
Supplementary Sec.~\ref{app:training_configurations} details optimization, screening, and the
shared training schedule.
Supplementary Sec.~\ref{app:curriculum_results} describes additional 4B configurations,
including the stage-ordered curriculum.

\paragraph{External Benchmarks.}
We evaluate transfer on MindCube~\cite{mindcube2025},
MMSI-Bench~\cite{mmsibench2026}, and SPBench~\cite{spatialladder2025},
with MMMU~\cite{mmmu2024} as a general multimodal reference.
SPBench contributes separate multiple-choice accuracy scores for
single-image (SI) and multi-view (MV) questions.
The aggregate transfer metric is the arithmetic mean of the five scores
before rounding.
Supplementary Sec.~\ref{app:external_results} provides further evaluation details.

\subsection{Benchmark Performance Across MLLMs}
\label{sec:main_results}
\label{sec:real_results}

\paragraph{Overall performance.}
We compare existing MLLMs to assess their ability to reason about
physical scene--projection relationships.
Sol and Gemini achieve 79.42\% and 57.71\% overall on the synthetic
test set, while the original open-weight models and existing spatially
trained baselines remain close to chance (Table~\ref{tab:main}).
This gap shows that the benchmark's physical projection questions
remain challenging for current open-weight spatial models.

\begin{table}[!t]
\caption{\textbf{SpatialCast and SpatialCast-Real performance.}
Accuracy (\%) on 4,500 synthetic questions (1,800 T1, 1,800 T2, 900 T3)
and 200 real-image questions.
Synthetic Overall is micro-averaged accuracy across
the three tasks.
Lavender marks SpatialCast-GRPO.
Bold marks each local block's maxima.}
\label{tab:main}
\centering
\footnotesize
\renewcommand{\arraystretch}{1.0}
\setlength{\tabcolsep}{1.5pt}
\begin{tabularx}{\columnwidth}{@{}>{\raggedright\arraybackslash}X*{5}{r}@{}}
\toprule
& \multicolumn{4}{c}{Synthetic (4,500)} & \multicolumn{1}{c}{Real (200)} \\
\cmidrule(lr){2-5}\cmidrule(l){6-6}
Model & T1 & T2 & T3 & Overall & Overall \\
\midrule
GPT-6.1 Sol & 67.50 & 88.39 & 85.33 & 79.42 & 79.50 \\
Gemini 4 Argon & 43.89 & 66.17 & 68.44 & 57.71 & 63.00 \\
\midrule
Qwen3-VL-8B Base~\cite{qwen3vl2025} & 22.89 & 25.44 & 25.33 & 24.40 & 34.50 \\
\rowcolor{ourslavender}
SpatialCast-GRPO (8B) & \textbf{30.94} & \textbf{48.78} & \textbf{34.22} & \textbf{38.73} & \textbf{36.50} \\
\midrule
Qwen3-VL-4B Base~\cite{qwen3vl2025} & 23.17 & 27.28 & 23.67 & 24.91 & 33.50 \\
Spatial-SSRL-4B~\cite{spatialssrl2026} & 23.33 & 28.56 & 23.33 & 25.42 & 36.50 \\
SpatialBlock-4B~\cite{spatialblock2026} & \textbf{24.22} & 24.83 & \textbf{25.56} & 24.73 & 34.00 \\
\rowcolor{ourslavender}
SpatialCast-GRPO (4B) & 24.11 & \textbf{46.83} & 22.44 & \textbf{32.87} & \textbf{38.50} \\
\midrule
Qwen2.5-VL-3B Base~\cite{qwen25vl2025} & 23.11 & 26.11 & 23.44 & 24.38 & 30.50 \\
Spatial-MLLM~\cite{spatialmllm} & 24.50 & 25.11 & \textbf{28.44} & 25.53 & 35.50 \\
SpatialLadder-3B~\cite{spatialladder2025} & 25.06 & 26.39 & 26.00 & 25.78 & 33.00 \\
Spatial-SSRL-3B~\cite{spatialssrl2026} & 25.11 & 27.72 & 23.89 & 25.91 & 34.50 \\
\rowcolor{ourslavender}
SpatialCast-GRPO (3B) & \textbf{26.83} & \textbf{40.17} & 25.22 & \textbf{31.84} & \textbf{36.00} \\
\midrule
InternVL3-2B Base~\cite{internvl32025} & 25.17 & 26.83 & 24.89 & 25.78 & 33.50 \\
\rowcolor{ourslavender}
SpatialCast-GRPO (2B) & \textbf{27.61} & \textbf{45.67} & \textbf{26.33} & \textbf{34.58} & \textbf{34.50} \\
\bottomrule
\end{tabularx}

\end{table}

\begin{table*}[!t]
\caption{\textbf{API-model accuracy across projection reasoning subtypes.}
Accuracy (\%) on the same 4,500 synthetic test questions, with column
groups and order matching Fig.~\ref{fig:subtype_gains}.
Dir./Len./Cont. denote shadow direction, length, and contour.
Azim./Ht. denote light azimuth and height.
L/R/C denote light, object-rotation, and camera viewpoint changes,
and paired letters denote joint interventions.}
\label{tab:api_subtype_comparison}
\centering
\footnotesize
\setlength{\tabcolsep}{2pt}
\renewcommand{\arraystretch}{1.12}
\begin{tabular*}{\textwidth}{@{\extracolsep{\fill}}l*{14}{r}@{}}
\toprule
& \multicolumn{5}{c}{T1 Compatibility} & \multicolumn{6}{c}{T2 Propagation} & \multicolumn{3}{c}{T3 Attribution} \\
\cmidrule(lr){2-6}\cmidrule(lr){7-12}\cmidrule(lr){13-15}
Model & Dir. & Len. & Cont. & Azim. & Ht. & L & R & C & LR & LC & RC & L & R & C \\
\midrule
GPT-6.1 Sol & 94.02 & 41.47 & 55.33 & 87.50 & 39.00 & 80.33 & 91.03 & 91.30 & 91.00 & 89.70 & 86.96 & 80.00 & 79.60 & 96.35 \\
Gemini 4 Argon & 60.47 & 27.76 & 33.33 & 55.33 & 31.00 & 65.67 & 64.12 & 65.22 & 85.00 & 61.13 & 55.85 & 78.33 & 56.52 & 70.43 \\
\bottomrule
\end{tabular*}

\par\vspace{4pt}
\centering
\includegraphics[width=0.95\textwidth]{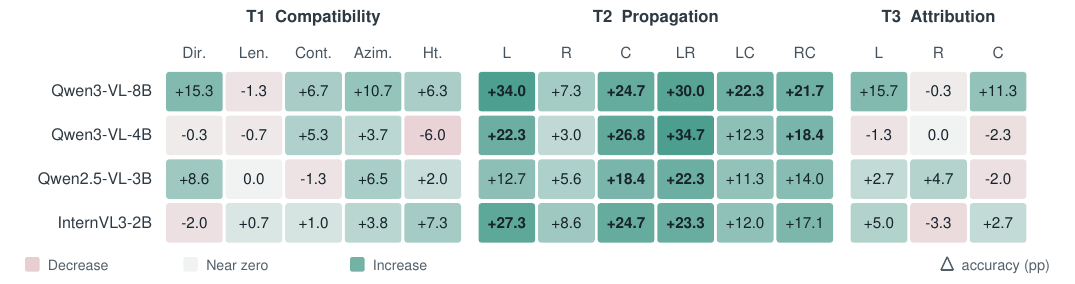}
\captionof{figure}{\textbf{Learning responses across projection reasoning subtypes.}
Signed accuracy changes after SpatialCast-GRPO relative to each model's
own Base ($\Delta$, percentage points), computed from integer correct
counts before rounding on the same 4,500-question test set.
Cell values
are rounded to one decimal.
The 8B row uses the final 1,399-update
checkpoint.
Rose denotes decreases, gray near-zero changes, and teal
increases.
Color intensity rises with absolute change.
Dir./Len./Cont. denote shadow direction, length, and contour.
Azim./Ht. denote light azimuth and height.
L/R/C denote light,
object-rotation, and camera changes.
Paired letters denote joint interventions.}
\label{fig:subtype_gains}
\end{table*}

\paragraph{API models across reasoning tasks.}
We examine how the differences between the API models vary across
reasoning tasks and physical factors (Table~\ref{tab:api_subtype_comparison}).
Both are weakest on T1, while Sol peaks on T2 and Gemini on T3.
Within T1, both perform better on shadow direction and light azimuth
than on shadow length and light height.
This shared pattern reveals a performance gap between directional
judgments and judgments of length or height in this benchmark.
Within T2, Sol varies less across intervention configurations, while
Gemini performs better on joint LR changes than on any single-factor
intervention.
Performance varies across intervention configurations, revealing distinct reasoning demands beyond the number of changed factors.
Within T3, Sol is strongest on camera-change attribution and Gemini
on light attribution.
The two models perform similarly on light attribution, while Sol's
advantages are substantially larger for object-rotation and camera-change
attribution.
Supplementary Sec.~\ref{app:gemini_analysis} gives subtype sample sizes and correct counts
by task and test partition.

\paragraph{Open-weight models and spatially trained baselines.}
Both the four base models and the spatially trained baselines remain
close to chance overall (Table~\ref{tab:main}).
The latter result characterizes transfer from existing spatial training
to these physical projection questions.
Section~\ref{sec:training_results} examines the improvements obtained
by training directly on \method.

\paragraph{API models on SpatialCast-Real.}
We further assess whether five API models can reason about photographed
lighting changes.
Their overall accuracy ranges from 58.50\% to 81.50\%, exceeding the
34.375\% random-choice expectation (Table~\ref{tab:real_reference}).
All models exceed the respective chance levels of 25\% and 50\% on
light-change propagation (T2) and attribution (T3), supporting the
answerability of these photographic questions for the API models.

\paragraph{Performance across evaluation settings.}
SpatialCast-trained models retain improvements on both scene-test and
geometry-test partitions (Supplementary Sec.~\ref{app:split_diagnostics}).
For example, the 8B model gains 13.37 and 16.27 percentage points,
respectively.
Its accuracy on SpatialCast-Real increases by 2.00 percentage points,
showing a smaller gain on real photographs than on the synthetic partitions.

\begin{table}[!t]
\caption{\textbf{API-model performance on SpatialCast-Real.}
Accuracy (\%) on 125 T2 and 75 T3 questions (200 overall).
Random denotes uniform guessing.
}
\label{tab:real_reference}
\centering
\footnotesize
\renewcommand{\arraystretch}{1.0}
\setlength{\tabcolsep}{3pt}
\begin{tabularx}{\columnwidth}{@{}>{\raggedright\arraybackslash}Xrrr@{}}
\toprule
Model & T2 & T3 & Overall \\
\midrule
Random & 25.00 & 50.00 & 34.38 \\
\midrule
GPT-6 Astra & 77.60 & 88.00 & 81.50 \\
GPT-6.1 Sol & 76.00 & 85.33 & 79.50 \\
Claude Sonnet 5.5 & 66.40 & 82.67 & 72.50 \\
Gemini 4 Argon & 52.80 & 80.00 & 63.00 \\
Qwen3.8-Max (0902)$^{\dagger}$ & 48.00 & 76.00 & 58.50 \\
\bottomrule
\end{tabularx}

\end{table}

\subsection{Learning Gains Across Reasoning Tasks}
\label{sec:training_results}

Figure~\ref{fig:subtype_gains} reveals distinct learning behaviors across
backbone models.
For example, shadow-direction compatibility (T1) improves by 15.3 percentage points
for 8B but remains nearly unchanged for 4B.
Similarly, light attribution (T3) gains 15.7 percentage points for 8B but decreases
by 1.3 percentage points for 4B.
Models improve on different reasoning subtypes even under the same training
setup.

Despite these differences, common trends emerge across models.
All four backbones improve on every Projection Propagation (T2) subtype,
whereas shadow-length compatibility (T1) shows minimal improvement and
object-rotation attribution (T3) exhibits limited or inconsistent gains.
These gains were measured under a training configuration that retained
all T2 training questions and selected subtypes of T1 and T3.

\subsection{Benchmark Validation}
\label{sec:diagnostics}
\label{sec:input_dependency}

\paragraph{Shadow Evidence Ablation.}
To examine whether MLLMs use cast shadows for reasoning, we conduct input
ablations on 240 light-change questions (T2-L), holding object geometry,
pose, and camera viewpoint fixed within each question.
\emph{Shadow-only} removes surface shading and floor illumination variation
while preserving cast shadows; \emph{No Shadow} removes cast shadows from
the same flat renderings; and \emph{No Initial} removes the initial
observation from Full Input
(Supplementary Sec.~\ref{app:input_dependency}).

\begin{table}[!t]
\caption{\textbf{Shadow evidence ablation on 240 T2-L questions.}
Original-label agreement (\%); shadow-erased scores are diagnostic.
See Supplementary Sec.~\ref{app:input_dependency}.}
\label{tab:input_dependency}
\centering
\footnotesize
\renewcommand{\arraystretch}{1.0}
\setlength{\tabcolsep}{2pt}
\begin{tabular*}{\columnwidth}{@{\extracolsep{\fill}}lrrrr@{}}
\toprule
Model & \shortstack{Full\\Input} & \shortstack{Shadow-\\only} & \shortstack{No\\Shadow} & \shortstack{No\\Initial} \\
\midrule
GPT-6.1 Sol & 81.3 & 81.7 & 25.0 & 41.7 \\
\midrule
Qwen3-VL-4B Base & 27.1 & 27.1 & 22.9 & 19.6 \\
SpatialCast-GRPO (4B) & 49.2 & 42.9 & 30.0 & 28.8 \\
\midrule
Qwen2.5-VL-3B Base & 26.3 & 25.4 & 22.9 & 20.0 \\
SpatialCast-GRPO (3B) & 34.6 & 35.0 & 25.4 & 24.6 \\
\midrule
InternVL3-2B Base & 24.2 & 27.1 & 24.6 & 24.2 \\
SpatialCast-GRPO (2B) & 51.3 & 52.1 & 23.3 & 25.0 \\
\bottomrule
\end{tabular*}

\end{table}

As shown in Table~\ref{tab:input_dependency}, Sol achieves 81.7\% under
Shadow-only, comparable to 81.3\% with full inputs, but drops to 25.0\%
without shadows.
The trained 4B, 3B, and 2B models show similar declines, from 42.9\%,
35.0\%, and 52.1\% to 30.0\%, 25.4\%, and 23.3\%, respectively.
These consistent declines suggest that cast shadows serve as important
visual evidence for reasoning about lighting changes.
Removing the initial observation also reduces performance for Sol and all
three GRPO models, indicating that the initial scene provides an important reference
for predicting the resulting projection.

\paragraph{Human answerability.}
Five participants achieve 65.00\% mean accuracy on a 32-question pilot, exceeding the 26.56\% random-choice expectation (Table~\ref{tab:pilot_reference}).
Every question receives at least one correct response, providing initial evidence of question answerability (Supplementary Sec.~\ref{app:human_pilot_subset}).

\begin{table}[!t]
\caption{\textbf{Human and API references on the 32-question pilot.}
Accuracy (\%).
Human mean averages five
participants.
Real denotes SpatialCast-Real.
Uniform random accounts
for two binary real-image T3 items.
All other items have four choices.}
\label{tab:pilot_reference}
\centering
\footnotesize
\renewcommand{\arraystretch}{1.0}
\setlength{\tabcolsep}{2.5pt}
\begin{tabular*}{\columnwidth}{@{\extracolsep{\fill}}lrrrrr@{}}
\toprule
32-item pilot & T1 & T2 & T3 & Real & Overall \\
\midrule
Human mean & 60.00 & 63.33 & 66.67 & 80.00 & 65.00 \\
GPT-6.1 Sol & 50.00 & 75.00 & 83.33 & 50.00 & 65.63 \\
Gemini 4 Argon & 50.00 & 66.67 & 50.00 & 75.00 & 59.38 \\
\midrule
Uniform random & 25.00 & 25.00 & 25.00 & 37.50 & 26.56 \\
\bottomrule
\end{tabular*}

\end{table}

\subsection{Cross-Benchmark Transfer}
\label{sec:transfer}

\begin{table}[!t]
\caption{\textbf{External transfer before and after SpatialCast training.}
Accuracy is reported as a percentage.
MC denotes MindCube, and SI/MV denote
SPBench single-/multi-view.
Avg. averages the five scores before rounding.
Bold marks the higher value in each
Base/trained pair, including ties.
Lavender marks SpatialCast-GRPO.
Results use greedy decoding. Full comparisons and numerical results appear
in Supplementary Sec.~\ref{app:external_results}.}
\label{tab:transfer}
\centering
\footnotesize
\renewcommand{\arraystretch}{1.0}
\setlength{\tabcolsep}{2pt}
\begin{tabularx}{\columnwidth}{@{}>{\raggedright\arraybackslash}X*{6}{r}@{}}
\toprule
Model & MC & MMSI & SI & MV & MMMU & Avg. \\
\midrule
Qwen3-VL-8B Base & 35.33 & 25.70 & 40.05 & \textbf{32.79} & \textbf{53.67} & 37.51 \\
\rowcolor{ourslavender}
SpatialCast-GRPO (8B) & \textbf{42.29} & \textbf{27.80} & \textbf{41.06} & \textbf{32.79} & 51.67 & \textbf{39.12} \\
\midrule
Qwen3-VL-4B Base & 33.81 & 25.00 & 38.54 & 31.15 & 52.00 & 36.10 \\
\rowcolor{ourslavender}
SpatialCast-GRPO (4B) & \textbf{40.38} & \textbf{27.00} & \textbf{42.07} & \textbf{39.34} & \textbf{52.89} & \textbf{40.34} \\
\midrule
Qwen2.5-VL-3B Base & 35.33 & \textbf{28.60} & 35.52 & \textbf{37.70} & \textbf{47.22} & 36.87 \\
\rowcolor{ourslavender}
SpatialCast-GRPO (3B) & \textbf{41.24} & 26.50 & \textbf{39.55} & 36.07 & 45.11 & \textbf{37.69} \\
\midrule
InternVL3-2B Base & 45.52 & \textbf{24.10} & 29.97 & \textbf{40.98} & \textbf{46.22} & \textbf{37.36} \\
\rowcolor{ourslavender}
SpatialCast-GRPO (2B) & \textbf{47.62} & 22.80 & \textbf{33.50} & 27.87 & 44.33 & 35.22 \\
\bottomrule
\end{tabularx}

\end{table}

\paragraph{External benchmark transfer.}
Table~\ref{tab:transfer} shows that SpatialCast training yields selective
improvements on external spatial benchmarks.
The average accuracy increases from 37.51\% to 39.12\% for 8B,
36.10\% to 40.34\% for 4B, and 36.87\% to 37.69\% for 3B.
For 2B, the five-metric average decreases from 37.36\% to 35.22\%.
All four models improve on MindCube and SPBench-SI.
On SPBench-MV, accuracy rises by 8.19 percentage points for 4B,
holds steady for 8B, and falls by 1.63 points for 3B and 13.11 points for 2B.
These results suggest that learning from physical projections can benefit
broader spatial reasoning, but the transfer remains model- and
task-dependent.

% ===== sec/5_conclusion =====
\section{Conclusion}
\label{sec:conclusion}

We introduced \method, a benchmark that uses cast shadows as a physical
readout of the 3D world to evaluate how MLLMs reason about the relationship
between scene conditions and visual observations.
By examining
compatibility, propagation, and attribution through a shared projection
process, \method reveals distinct reasoning capabilities that conventional
aggregate evaluations can obscure.
Our experiments show substantial
differences across models and reasoning directions, with GRPO-based
training consistently producing the largest gains in propagation across
three model families.
These improvements extend to held-out scenes and
geometric conditions, while evaluations on real images and external
benchmarks reveal varying degrees of generalization.
More broadly, \method treats shadows not merely as visual patterns to
recognize, but as physical evidence for probing whether MLLMs can reason
about the causes and consequences of changes in the 3D world.

\clearpage

{\small
\bibliographystyle{ieeenat_fullname}
\bibliography{main}}

\clearpage

\appendix
\raggedbottom
\onecolumn

% Supplementary heading
\section*{Supplementary Material}
% Only collect supplementary sections
\startcontents[supplementary]

\pdfbookmark[0]{Supplementary Contents}{supp:contents}

\begin{multicols}{2}[\section*{Contents}]
\printcontents[supplementary]{}{1}{%
  \setcounter{tocdepth}{2}%
}
\end{multicols}

% ===== sec/7_appendix =====
\section{Synthetic Benchmark Details}
\label{app:data}

\subsection{Shape Families and Coverage}
\label{app:a1}
Across the synthetic pool, each of the 16 families in
Figure~\ref{fig:shape_gallery} appears in 1,757--2,461 questions before
training selection. A question with two different families counts toward both.

\noindent\begin{minipage}{\textwidth}
    \centering
    \captionsetup{hypcap=false}
    \includegraphics[width=0.75\textwidth]{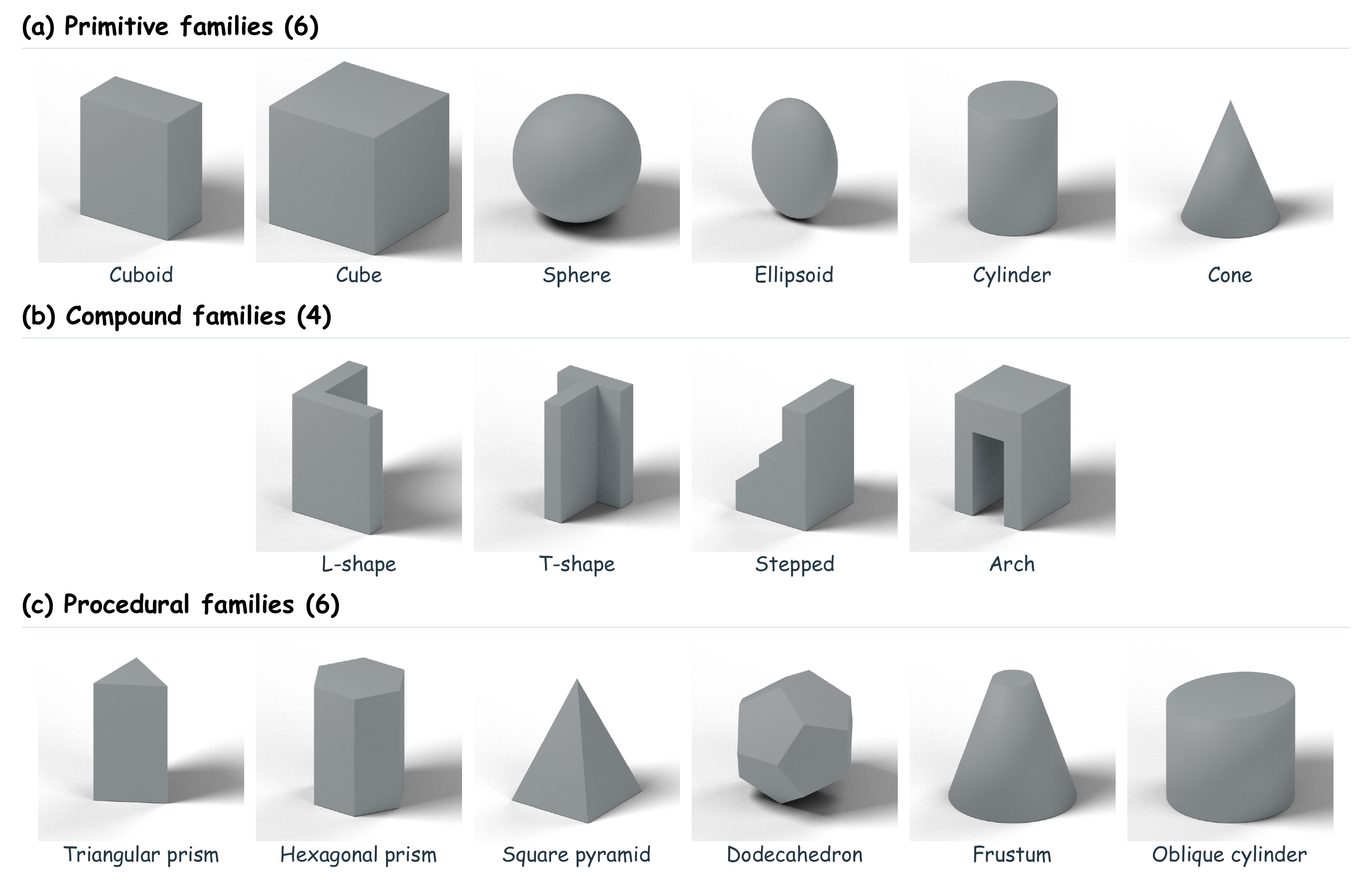}
    \captionof{figure}{\textbf{All 16 synthetic shape families.}
    (a) Six primitive, (b) four compound, and (c) six procedural families.
    Each panel renders an accepted dataset geometry with its original
    mesh parts and proportions. Gallery views use a common neutral material,
    lighting, orthographic camera, and pose, with uniform per-mesh scaling
    for display.}
    \label{fig:shape_gallery}
\end{minipage}

\subsection{Supplementary T2 Examples}
\label{app:question_examples}
Figure~\ref{fig:t2_single_examples} shows all four candidates for the main
paper's light ($L$), object rotation ($R$), and camera viewpoint ($C$)
examples. Figure~\ref{fig:t2_combined_examples} shows the joint interventions
$LR$, $LC$, and $RC$. Both figures preserve the original candidate order.

\twocolumn[{
\noindent\begin{minipage}{\textwidth}
    \captionsetup{hypcap=false}
    \centering
    \includegraphics[width=\textwidth]{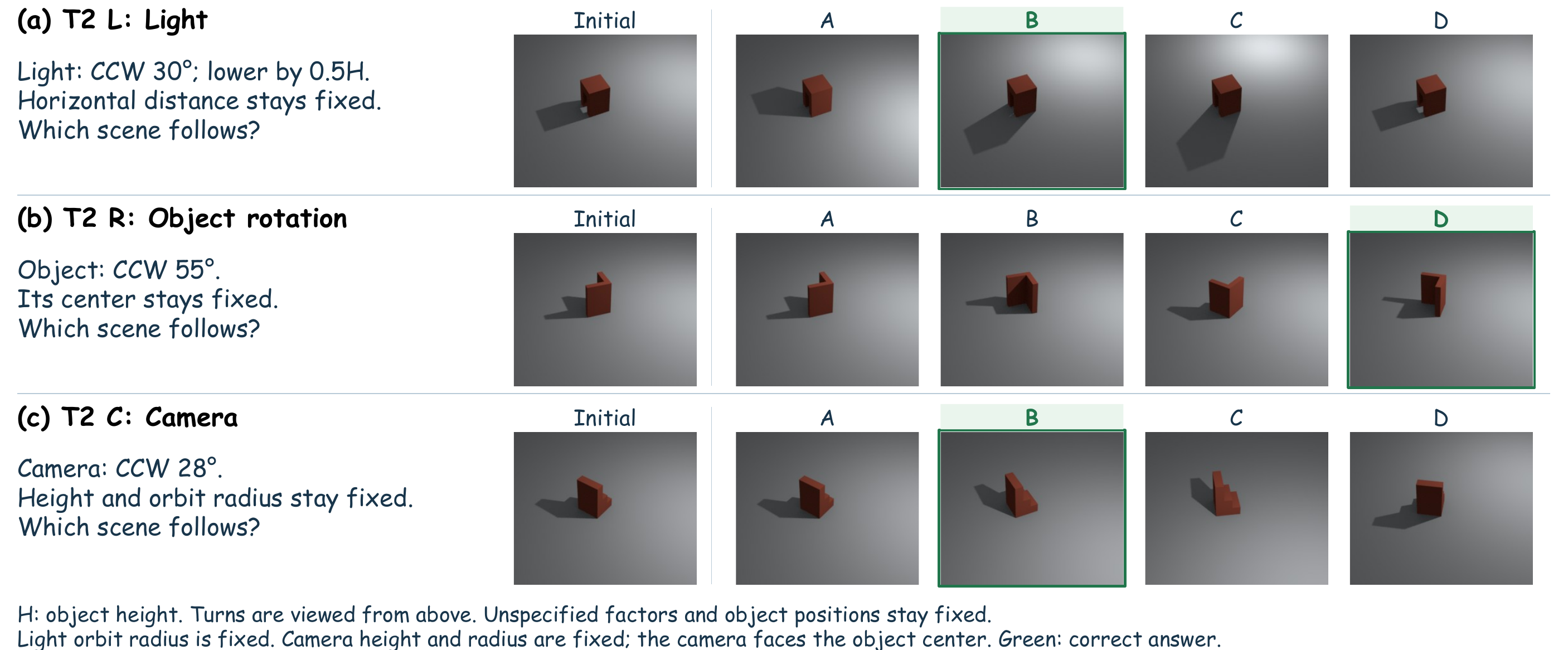}
    \captionof{figure}{\textbf{Single-factor intervention questions with complete
    candidate sets.} Each row
    gives the specified intervention, the initial observation, and four
    candidate outcomes. Green marks the correct candidate.}
    \label{fig:t2_single_examples}
\par\vspace{10pt}
    \centering
    \includegraphics[width=\textwidth]{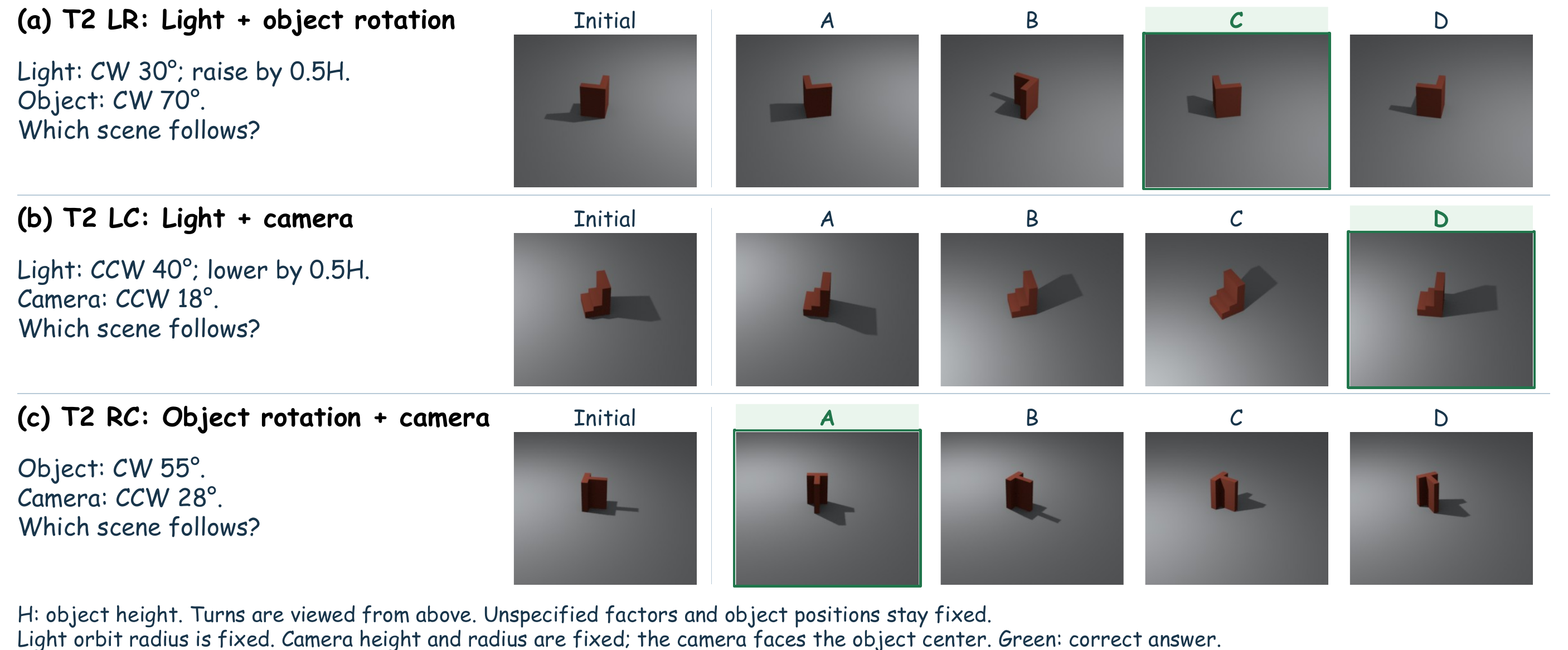}
    \captionof{figure}{\textbf{Joint intervention questions.} The three rows
    illustrate light plus object rotation ($LR$), light plus camera
    ($LC$), and object rotation plus camera ($RC$). The correct outcome
    must satisfy both specified changes. Green marks the correct
    candidate.
    The remaining candidates include omitted or reversed
    intervention components.}
    \label{fig:t2_combined_examples}
\end{minipage}
\vspace{5pt}
}]

\subsection{Dataset Splits and Complete Counts}
\label{app:split}
Scene-test uses independently generated instances from the training sampler.
Geometry-test uses disjoint, shape-specific geometric parameter ranges within
13 shape families. Spheres, cubes, and regular dodecahedra are excluded
because no geometry-holdout parameter is defined for them. Both objects in
two-object scenes must satisfy the held-out ranges.
For example, the revised generator samples the cuboid height-to-maximum-base-extent
ratio from $[1.2,1.6]$ in-domain and $[1.9,2.3]$ for geometry-test.
Table~\ref{tab:split_counts} gives the counts for each subtype. T1 forward
questions ask about shadow contour, direction, or length. Inverse
questions ask about light azimuth or height.

\begin{table}[!htbp]
\caption{\textbf{Synthetic split counts before training selection.}
Each subtype covers single objects and object pairs with separate or
overlapping shadows. Scene and Geometry denote test splits.}
\label{tab:split_counts}
\centering\footnotesize
\renewcommand{\arraystretch}{0.94}
\setlength{\tabcolsep}{1.5pt}
\begin{tabular*}{\columnwidth}{@{\extracolsep{\fill}}lrrrrr@{}}
\toprule
Subtype & Train & Val & Scene & Geometry & Total \\
\midrule
T1 Contour & 1,002 & 98 & 201 & 99 & 1,400 \\
T1 Direction & 998 & 100 & 201 & 100 & 1,399 \\
T1 Length & 1,000 & 102 & 198 & 101 & 1,401 \\
T1 Light azimuth & 1,998 & 198 & 402 & 198 & 2,796 \\
T1 Light height & 1,002 & 102 & 198 & 102 & 1,404 \\
\midrule
T2 L & 999 & 100 & 199 & 101 & 1,399 \\
T2 R & 1,000 & 102 & 203 & 98 & 1,403 \\
T2 C & 1,001 & 98 & 198 & 101 & 1,398 \\
T2 LR & 998 & 99 & 199 & 101 & 1,397 \\
T2 LC & 1,002 & 101 & 201 & 100 & 1,404 \\
T2 RC & 1,000 & 100 & 200 & 99 & 1,399 \\
\midrule
T3 L & 998 & 101 & 201 & 99 & 1,399 \\
T3 R & 1,001 & 100 & 198 & 101 & 1,400 \\
T3 C & 1,001 & 99 & 201 & 100 & 1,401 \\
\midrule
Total & 15,000 & 1,500 & 3,000 & 1,500 & 21,000 \\
\bottomrule
\end{tabular*}

\end{table}

\subsection{Quality Checks}
\label{app:quality}
Across the 21,000 synthetic questions, each question ID occurs in exactly
one of the four partitions in Table~\ref{tab:split_counts}.

We assess visual separation using the same criterion across all four
partitions. For each of the six pairs of candidate scene renders, we
count pixels in the object or shadow region where at least one 8-bit
RGB channel differs by 12 or more. In 20,552 of 21,000 questions
(97.87\%), even the closest pair differs at 150 or more such pixels.
See Appendix~\ref{app:training_configurations} for training selection.

\section{SpatialCast-Real Construction}
\label{app:real_construction}
\begin{figure*}[!t]
    \centering
    \includegraphics[width=0.88\textwidth]{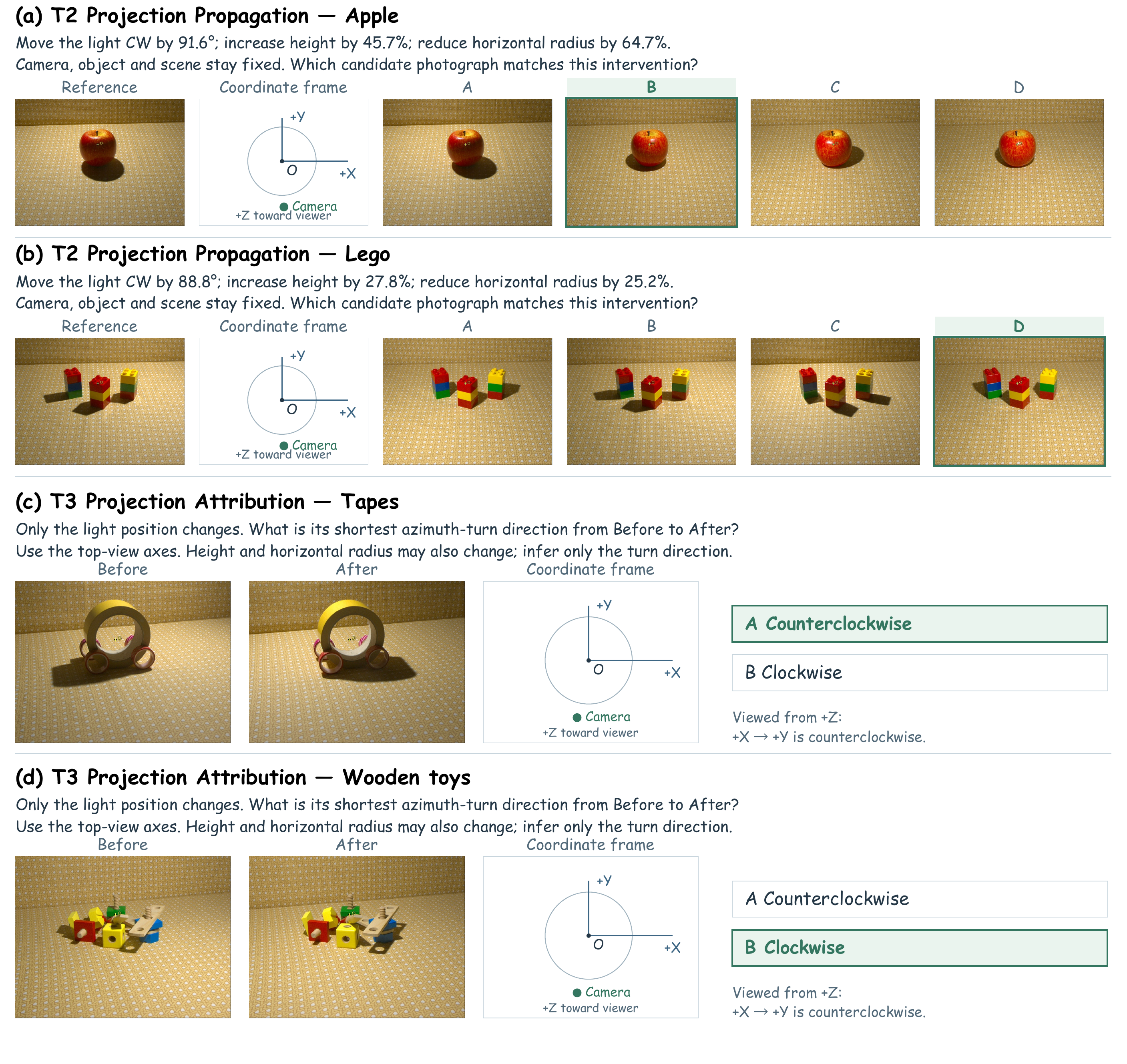}
    \caption{\textbf{Complete SpatialCast-Real questions from ReNe.}
    Panels (a,b) show T2 candidate photographs in their original order.
    Panels (c,d) show T3 before/after pairs. Green marks the correct answer.
    CW denotes clockwise as viewed from above along $-Z$. The top views
    show the world axes and calibrated camera bearing.}
    \label{fig:real_question_examples}
\end{figure*}

\subsection{Source Data and Ground-Truth Derivation}
We use single-light photographs from all 20 ReNe scenes~\cite{toschi2023relight}.
Each scene contributes five questions from camera 00 and five from camera 20,
giving 100 questions per camera and 200 overall (125 T2 and 75 T3).
The questions use 585 distinct photographs and two top-view diagrams.
Within each question, the scene and camera are fixed while the light moves.

In ReNe's world coordinates, $O$ is the origin and $+Z$ points upward.
For a calibrated light position $(x,y,z)$, the azimuth is
$\theta=\operatorname{atan2}(y,x)$, the height is $h=z$, and the
horizontal radius is $\rho=\sqrt{x^2+y^2}$. Azimuth increases
counterclockwise from $+X$ toward $+Y$ when viewed from above.
For a reference and target photograph, we compute the shortest azimuth
turn and the percentage changes in height and radius as follows.
\begin{equation}
\begin{aligned}
\Delta\theta &= \operatorname{wrap}_{[-180^\circ,180^\circ)}
 (\theta_{\mathrm{target}}-\theta_{\mathrm{ref}}),\\
\Delta h_{\%} &=100(h_{\mathrm{target}}/h_{\mathrm{ref}}-1),\\
\Delta\rho_{\%} &=100(\rho_{\mathrm{target}}/\rho_{\mathrm{ref}}-1).
\end{aligned}
\label{eq:real_light_change}
\end{equation}
For T2, Eq.~\ref{eq:real_light_change} supplies the three changes in the
prompt. The photograph recorded at the target light position is the
correct candidate. For T3, the sign of $\Delta\theta$ determines the turn direction. Positive values indicate
counterclockwise turns and negative values indicate clockwise turns. Height and
radius may also change between the two photographs.

\subsection{Question Examples and Quality Control}
Figure~\ref{fig:real_question_examples} shows two complete examples of
each task. Across the dataset, the correct T2 option appears in positions
A/B/C/D 32/31/31/31 times. T3 has 38 counterclockwise and 37 clockwise turns.

We check all six candidate pairs in each T2 question (750 pairs overall).
The minimum azimuth separation is $45^\circ$. Within a fixed region of
interest on $256\!\times\!192$ previews, each pair must have a normalized
mean absolute difference of at least 0.10, with at least 25\% of pixels
differing by 0.12 or more. Blockwise structural similarity (block-SSIM)
must be at most 0.82. We retain the light and camera records,
question IDs, candidate mappings, and image hashes for reconstruction.

\addtocontents{toc}{\protect\columnbreak}
\section{Training and Evaluation Details}
\label{app:training}

\subsection{Data Selection and GRPO Optimization}
\label{app:training_configurations}

We retain 11,186 of 15,000 training questions, comprising 1,189 forward T1 questions
screened for visually distinct candidates, 1,998 inverse T1 questions on
light azimuth, all 6,000 T2 questions, and 1,999 T3 questions on light or
camera changes. All four backbones use the same shuffle (seed 1) for
1,399 updates of eight questions, repeating six questions to fill the last batch.

GRPO samples eight responses per question and rewards correctness (0.9)
and format compliance (0.1). For correct responses, the length penalty
rises linearly from zero at 1,536 tokens to 0.1 at 2,048 tokens.
All four backbones freeze the visual tower and use AdamW (constant learning
rate $10^{-6}$, weight decay 0.01), rollout temperature 1.0, a
2,048-token limit, KL coefficient 0.01, and clipping ratios 0.2/0.3.
Appendix D.2.2 compares the additional 4B configurations.

\subsection{Model Evaluation and Answer Parsing}
\label{app:small_model_evaluation}
\label{app:api_evaluation}

\paragraph{Local inference.}
For the 2B, 3B, 4B, and 8B comparisons, the base and trained checkpoints of
each backbone use the same questions, prompts, and image processing.
We use greedy decoding with a 2,048-token output limit. InternVL3-2B
uses its native image processor and chat template with a direct-answer
prompt. Accuracy counts unresolved or invalid answers as incorrect.

\paragraph{API inference.}
We evaluate all five API models on
the 200 SpatialCast-Real questions, and GPT-6.1 Sol and Gemini 4 Argon
on the synthetic test. OpenRouter requests include original-resolution
images and ask for the option letter. We use medium reasoning effort,
an 8,192-token output limit, and omit the temperature parameter.
The model identifiers are \texttt{openai/gpt-6-astra},
\texttt{openai/gpt-6.1-sol}, \texttt{anthropic/claude-sonnet-5.5},
\texttt{google/gemini-4-argon}, and
\texttt{qwen/qwen3.8-max-0902}. The synthetic prompts interleave images
at their placeholders and retain option labels (e.g., A--D), without
revealing the correct answers.

\paragraph{Answer parsing.}
One deterministic parser scores all five API models. It first checks the
entire response for an allowed option letter or exact option text, then
checks the last nonempty line if needed. It accepts optional brackets,
terminal punctuation, balanced Markdown emphasis, an answer prefix, or
a complete \texttt{<answer>} tag. Each question is scored once.
For repeated requests, we use the last response.
Appendix~\ref{app:score_aggregation} gives the aggregation rule and correct counts.

\section{Supplementary Results and Metrics}
\label{app:results}

\subsection{Complete Benchmark Results}
\subsubsection{API Synthetic Evaluation}
\label{app:gemini_analysis}
\label{app:sol_analysis}
Table~\ref{tab:api_synthetic_counts} gives correct counts by task, task
group, and test partition for the same 4,500 questions. Its caption gives
the sample sizes for the fourteen subtype accuracies in the main paper.
All questions enter the denominator, including four invalid final choices
for Gemini and two empty responses for Sol.

\begin{table}[!htbp]
\caption{\textbf{API synthetic correct counts by task and partition.}
Split rows also report accuracy. Subtype sizes in the
main comparison are 301/299/300/600/300 for T1 direction/length/contour/
azimuth/height, 300/301/299/300/301/299 for T2 L/R/C/LR/LC/RC,
and 300/299/301 for T3 L/R/C.}
\label{tab:api_synthetic_counts}
\centering\footnotesize
\setlength{\tabcolsep}{3pt}
\begin{tabular*}{\columnwidth}{@{\extracolsep{\fill}}lrr@{}}
\toprule
Task / split & GPT-6.1 Sol & Gemini 4 Argon \\
\midrule
T1 (1,800) & 1,215 & 790 \\
T2 (1,800) & 1,591 & 1,191 \\
T3 (900) & 768 & 616 \\
Total (4,500) & 3,574 & 2,597 \\
\midrule
T1 forward (900) & 573 & 365 \\
T1 inverse (900) & 642 & 425 \\
T2 single-factor (900) & 788 & 585 \\
T2 joint-factor (900) & 803 & 606 \\
\midrule
Scene-test (3,000) & 2,373 (79.10\%) & 1,716 (57.20\%) \\
Geometry-test (1,500) & 1,201 (80.07\%) & 881 (58.73\%) \\
\bottomrule
\end{tabular*}

\end{table}

\subsubsection{Local-Model SpatialCast-Real Results}
\label{app:score_aggregation}
SpatialCast-Real contains 125 T2 and 75 T3 questions. Its overall accuracy is
$100(C_{\mathrm{T2}}+C_{\mathrm{T3}})/200$, where $C_{\mathrm{T2}}$ and
$C_{\mathrm{T3}}$ count correct T2 and T3 answers, respectively.
Table~\ref{tab:real_local} reports the local models' task scores.
The T2/T3 correct counts are 97/66 for GPT-6 Astra, 95/64 for GPT-6.1
Sol, 83/62 for Claude Sonnet 5.5, 66/60 for Gemini 4 Argon, and
60/57 for Qwen3.8-Max (0902).

\begin{table}[!htbp]
\caption{\textbf{Local-model performance on SpatialCast-Real.}
Accuracy (\%) on 125 T2 and 75 T3 questions. Overall covers all 200.
Random is the uniform-choice expectation. Blue marks SpatialCast-GRPO.
Each trained model is compared with its own Base.}
\label{tab:real_local}
\centering
\footnotesize
\setlength{\tabcolsep}{3pt}
\begin{tabularx}{\columnwidth}{@{}>{\raggedright\arraybackslash}Xrrr@{}}
\toprule
Model & \shortstack[r]{T2\\(125)} & \shortstack[r]{T3\\(75)} & \shortstack[r]{Overall\\(200)} \\
\midrule
Random & 25.00 & 50.00 & 34.38 \\
\midrule
\multicolumn{4}{@{}l}{\emph{Qwen3-VL-8B}} \\
Qwen3-VL-8B Base~\cite{qwen3vl2025} & 25.60 & 49.33 & 34.50 \\
\rowcolor{oursblue}
SpatialCast-GRPO (8B) & 28.00 & 50.67 & 36.50 \\
\midrule
\multicolumn{4}{@{}l}{\emph{Qwen3-VL-4B}} \\
Qwen3-VL-4B Base~\cite{qwen3vl2025} & 24.00 & 49.33 & 33.50 \\
Spatial-SSRL-4B~\cite{spatialssrl2026} & 28.00 & 50.67 & 36.50 \\
SpatialBlock-4B~\cite{spatialblock2026} & 24.00 & 50.67 & 34.00 \\
\rowcolor{oursblue}
SpatialCast-GRPO (4B) & 30.40 & 52.00 & 38.50 \\
\midrule
\multicolumn{4}{@{}l}{\emph{Qwen2.5-VL-3B}} \\
Qwen2.5-VL-3B Base~\cite{qwen25vl2025} & 24.80 & 40.00 & 30.50 \\
Spatial-MLLM~\cite{spatialmllm} & 26.40 & 50.67 & 35.50 \\
SpatialLadder-3B~\cite{spatialladder2025} & 22.40 & 50.67 & 33.00 \\
Spatial-SSRL-3B~\cite{spatialssrl2026} & 24.80 & 50.67 & 34.50 \\
\rowcolor{oursblue}
SpatialCast-GRPO (3B) & 28.00 & 49.33 & 36.00 \\
\midrule
\multicolumn{4}{@{}l}{\emph{InternVL3-2B}} \\
InternVL3-2B Base & 23.20 & 50.67 & 33.50 \\
\rowcolor{oursblue}
SpatialCast-GRPO (2B) & 24.80 & 50.67 & 34.50 \\
\bottomrule
\end{tabularx}

\end{table}

\subsection{Learning Responses and Training Configurations}
\subsubsection{Subtype Gains and Comparison Scope}
\label{app:learning_scope}

Table~\ref{tab:task_counts} gives the correct counts for each backbone
before and after training. For example, Qwen3-VL-8B gains 420 correct T2
answers on 1,800 questions (23.33 percentage points). Across all four
backbones, accuracy increases on each of the six T2 subtypes.
The shared training mixture includes all 6,000 T2 questions (53.64\%
of the 11,186 retained items) and selected T1 and T3 subtypes.
The gains therefore describe learning under this training distribution.

\begin{table}[!htbp]
\caption{\textbf{Exact synthetic correct counts.} Trained denotes
each backbone's final SpatialCast-GRPO checkpoint.}
\label{tab:task_counts}
\centering\footnotesize
\setlength{\tabcolsep}{3pt}
\begin{tabular*}{\columnwidth}{@{\extracolsep{\fill}}lrrrr@{}}
\toprule
Model pair & T1 & T2 & T3 & Total \\
Items & 1,800 & 1,800 & 900 & 4,500 \\
\midrule
8B Base & 412 & 458 & 228 & 1,098 \\
8B trained & 557 & 878 & 308 & 1,743 \\
4B Base & 417 & 491 & 213 & 1,121 \\
4B trained & 434 & 843 & 202 & 1,479 \\
3B Base & 416 & 470 & 211 & 1,097 \\
3B trained & 483 & 723 & 227 & 1,433 \\
2B Base & 453 & 483 & 224 & 1,160 \\
2B trained & 497 & 822 & 237 & 1,556 \\
\bottomrule
\end{tabular*}

\end{table}

\subsubsection{Additional 4B Training Configurations}
\label{app:curriculum_results}

\begin{table}[!htbp]
\caption{\textbf{Post-training configurations on the same 4B base.}
All configurations use the same test sets and direct-answer protocol.
The question count gives the number of distinct training items.
Bold marks column maxima. Both panels list the same 4B configurations;
Real denotes SpatialCast-Real.}
\label{tab:training_configs}
\centering
\footnotesize
\renewcommand{\arraystretch}{1.0}
\setlength{\tabcolsep}{2pt}
\begin{tabular*}{\columnwidth}{@{\extracolsep{\fill}}lrr@{}}
\toprule
4B configuration & Questions & Steps \\
\midrule
Qwen3-VL-4B Base & -- & -- \\
Curriculum & 15,000 & 1,875 \\
Full shuffle & 15,000 & 1,875 \\
SpatialCast-GRPO (4B) & 11,186 & 1,399 \\
T2-only & 6,000 & 750 \\
\bottomrule
\end{tabular*}
\par\medskip
\begin{tabular*}{\columnwidth}{@{\extracolsep{\fill}}lrrrrr@{}}
\toprule
& \multicolumn{4}{c}{Synthetic test (4,500 questions)} & Real \\
\cmidrule(lr){2-5}\cmidrule(l){6-6}
4B configuration & T1 & T2 & T3 & Overall & Overall \\
\midrule
Qwen3-VL-4B Base & 23.17 & 27.28 & 23.67 & 24.91 & 33.50 \\
Curriculum & 23.83 & 42.17 & 23.33 & 31.07 & 35.00 \\
Full shuffle & \textbf{24.39} & 43.06 & \textbf{25.89} & 32.16 & 36.00 \\
SpatialCast-GRPO (4B) & 24.11 & \textbf{46.83} & 22.44 & \textbf{32.87} & 38.50 \\
T2-only & 22.83 & 39.06 & \textbf{25.89} & 29.93 & \textbf{39.00} \\
\bottomrule
\end{tabular*}
\end{table}

Across the four runs in Table~\ref{tab:training_configs}, T2 shows the
largest synthetic task gain over the 4B base. The curriculum processes
T1, T2, and T3 in sequence, preserving within-task order and ending
the stages at updates 750, 1,500, and 1,875. Both curriculum and full
shuffle use 15,000 questions and 1,875 updates. Comparisons with
SpatialCast-GRPO also vary the training mixture and update budget.

\subsection{Cast-Shadow Ablation Settings}
\label{app:input_dependency}

\begin{figure}[!htbp]
\centering
{\footnotesize
\setlength{\tabcolsep}{0pt}
\begin{tabular}{@{}*{5}{>{\centering\arraybackslash}p{0.2\columnwidth}}@{}}
Initial & A & B & C & D \\
\end{tabular}\par
\smallskip
\includegraphics[width=\columnwidth,trim=0bp 244bp 0bp 0bp,clip]{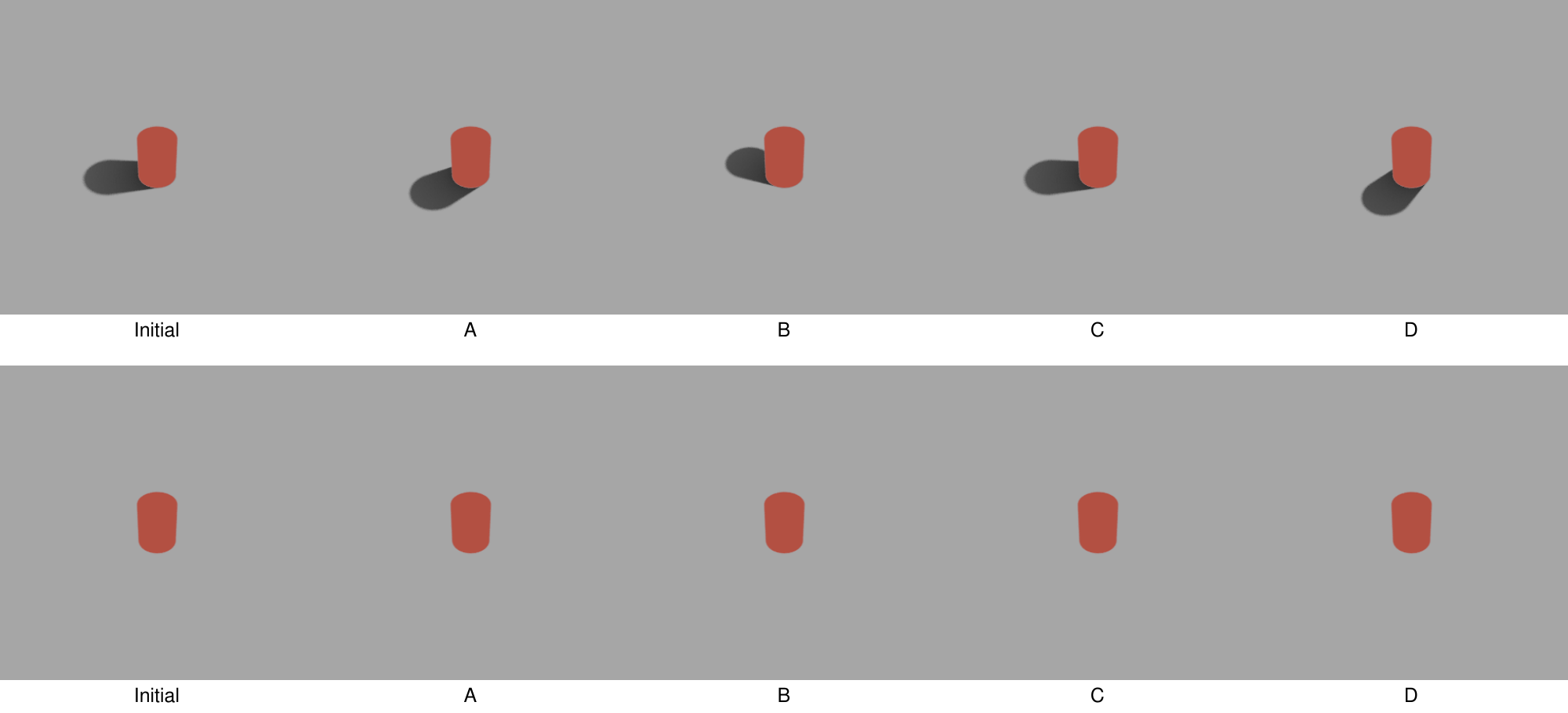}\par
\smallskip
\includegraphics[width=\columnwidth,trim=0bp 20bp 0bp 224bp,clip]{supplementary/flat_shadow_control.pdf}\par
}
\caption{\textbf{Shadow-only and its matched shadow-erased control.}
A T2-L example is shown in initial/A--D order. Shadow-only (upper row)
preserves cast-shadow differences on fixed object and floor colors.
No Shadow (lower row) removes these differences, making all five
images identical.}
\label{fig:controlled_shadow_evidence}
\end{figure}

\paragraph{Evaluation set.}
\label{app:input_dependency_protocol}
All conditions use the same 240 T2-L questions, original answer labels,
and candidate order, with inference settings held fixed for each model.

\paragraph{Ablation conditions.}
\label{app:visual_ablations}
\emph{Shadow-only} replaces object appearance with uniform colors and
modulates a constant floor color by the original cast-shadow attenuation.
It retains object masks and shadow geometry while removing surface shading
and floor illumination variation. \emph{No Shadow} uses the same
colors, masks, and compositing procedure with shadow attenuation removed;
the two conditions differ only in the retained cast shadows. Both edits
apply to the initial image and all four candidates. \emph{No Initial}
removes the initial image and its prompt marker from Full Input while
retaining the intervention description and all four candidate images.

\paragraph{Interpretation.}
\label{app:controlled_shadow_evidence}
Removing shadows can make candidates visually identical
(Fig.~\ref{fig:controlled_shadow_evidence}); No Shadow therefore reports
original-label agreement after evidence removal, not accuracy on uniquely
answerable questions. The matched Shadow-only versus No Shadow comparison
isolates cast-shadow evidence, whereas Full Input versus Shadow-only also
changes surface shading and floor illumination.

\subsection{Human Performance on a 32-Question Pilot}
\label{app:human_pilot_subset}

Five participants answered the same 32 questions against a fixed key.
The booklet pairs one single-object and one two-object scene-test question
for each of the fourteen synthetic subtypes, plus two T2 and two T3
questions from SpatialCast-Real. Participants answered 104/160 correctly
(65.00\%, compared with a 26.56\% uniform-choice expectation), and each
of the 32 questions received at least one correct response
(Table~\ref{tab:human_pilot32}). These results provide initial evidence
of answerability on this booklet.

\begin{table}[!htbp]
\caption{\textbf{Five participants on the 32-question pilot.}
Accuracy (\%). Item counts apply to each participant. Human mean
aggregates 104 correct responses across five participants answering
the same 32 items. Random is the uniform-choice expectation,
accounting for the two binary real-image T3 questions.}
\label{tab:human_pilot32}
\centering
\footnotesize
\setlength{\tabcolsep}{3pt}
\begin{tabular*}{\columnwidth}{@{\extracolsep{\fill}}lrrrrr@{}}
\toprule
& T1 & T2 & T3 & Real & Overall \\
Items & 10 & 12 & 6 & 4 & 32 \\
\midrule
P1 & 50.00 & 83.33 & 33.33 & 75.00 & 62.50 \\
P2 & 60.00 & 58.33 & 50.00 & 100.00 & 62.50 \\
P3 & 70.00 & 58.33 & 83.33 & 75.00 & 68.75 \\
P4 & 70.00 & 66.67 & 100.00 & 75.00 & 75.00 \\
P5 & 50.00 & 50.00 & 66.67 & 75.00 & 56.25 \\
Human mean & 60.00 & 63.33 & 66.67 & 80.00 & 65.00 \\
\midrule
Random & 25.00 & 25.00 & 25.00 & 37.50 & 26.56 \\
\bottomrule
\end{tabular*}

\end{table}

\subsection{Performance on Held-Out Test Partitions}
\label{app:split_diagnostics}
All four backbones gain accuracy on both held-out partitions
(Table~\ref{tab:diagnostics}). For each checkpoint, the two split counts
sum to the synthetic total in Table~\ref{tab:task_counts}.

\begin{table}[!htbp]
\caption{\textbf{Correct answers before and after SpatialCast training.}
Each entry gives base $\to$ trained counts on the same scene-test
(3,000 questions) or geometry-test (1,500 questions) partition.}
\label{tab:diagnostics}
\centering\footnotesize
\setlength{\tabcolsep}{2pt}
\begin{tabular*}{\columnwidth}{@{\extracolsep{\fill}}lrr@{}}
\toprule
Model & Scene-test & Geometry-test \\
\midrule
Qwen3-VL-8B   & $738\to1{,}139$ & $360\to604$ \\
Qwen3-VL-4B   & $751\to1{,}004$ & $370\to475$ \\
Qwen2.5-VL-3B & $726\to951$     & $371\to482$ \\
InternVL3-2B  & $779\to1{,}040$ & $381\to516$ \\
\bottomrule
\end{tabular*}

\end{table}

% ===== complete external benchmark comparison =====
\begin{table}[!htbp]
\caption{\textbf{Complete external evaluation (\%).} MC denotes MindCube. SPBench SI/MV report single-/multi-view multiple-choice
accuracy. Avg.\ averages the five scores before rounding. Bold marks each block's best
values, including ties. All reported GRPO-trained models use SpatialCast-GRPO.
Numerical-question results appear in Table~\ref{tab:numerical}.}
\label{tab:external_full}
\centering
\footnotesize
\renewcommand{\arraystretch}{1.0}
\setlength{\tabcolsep}{1.2pt}
\begin{tabular*}{\columnwidth}{@{\extracolsep{\fill}}lrrrrrr@{}}
\toprule
& & & \multicolumn{2}{c}{SPBench} & & \\
\cmidrule(lr){4-5}
Model & MC & MMSI & SI & MV & MMMU & Avg. \\
\midrule
\multicolumn{7}{l}{\emph{Qwen3-VL-8B}} \\
Base~\cite{qwen3vl2025} & 35.33 & 25.70 & 40.05 & \textbf{32.79} & \textbf{53.67} & 37.51 \\
SpatialCast-GRPO & \textbf{42.29} & \textbf{27.80} & \textbf{41.06} & \textbf{32.79} & 51.67 & \textbf{39.12} \\
\midrule
\multicolumn{7}{l}{\emph{Qwen3-VL-4B}} \\
Base~\cite{qwen3vl2025} & 33.81 & 25.00 & 38.54 & 31.15 & 52.00 & 36.10 \\
Spatial-SSRL~\cite{spatialssrl2026} & 28.29 & \textbf{27.50} & 37.78 & 32.79 & 52.67 & 35.81 \\
SpatialBlock~\cite{spatialblock2026} & \textbf{49.24} & 24.00 & 41.56 & 36.07 & 47.11 & 39.60 \\
SpatialCast-GRPO & 40.38 & 27.00 & \textbf{42.07} & \textbf{39.34} & \textbf{52.89} & \textbf{40.34} \\
\midrule
\multicolumn{7}{l}{\emph{Qwen2.5-VL-3B}} \\
Base~\cite{qwen25vl2025} & 35.33 & \textbf{28.60} & 35.52 & 37.70 & \textbf{47.22} & 36.87 \\
SpatialBlock~\cite{spatialblock2026} & \textbf{50.29} & 26.10 & 35.52 & 36.07 & 43.33 & 38.26 \\
Spatial-MLLM~\cite{spatialmllm} & 37.24 & 23.60 & 34.76 & 47.54 & 26.67 & 33.96 \\
SpatialLadder~\cite{spatialladder2025} & 47.81 & 26.50 & \textbf{52.14} & \textbf{54.10} & 44.89 & \textbf{45.09} \\
Spatial-SSRL~\cite{spatialssrl2026} & 31.05 & 24.90 & 35.52 & 36.07 & \textbf{47.22} & 34.95 \\
SpatialCast-GRPO & 41.24 & 26.50 & 39.55 & 36.07 & 45.11 & 37.69 \\
\midrule
\multicolumn{7}{l}{\emph{InternVL3-2B}} \\
Base~\cite{internvl32025} & 45.52 & \textbf{24.10} & 29.97 & \textbf{40.98} & \textbf{46.22} & \textbf{37.36} \\
SpatialCast-GRPO & \textbf{47.62} & 22.80 & \textbf{33.50} & 27.87 & 44.33 & 35.22 \\
\bottomrule
\end{tabular*}

\end{table}

\subsection{External Benchmark Results}
\label{app:external_results}
\label{app:numerical}

\subsubsection{Complete Greedy Model Comparison}
Table~\ref{tab:external_full} reports greedy evaluation on five external
metrics. The average weights the five metrics equally. Relative to their
own bases, all four trained backbones improve on MindCube and SPBench
single-image multiple choice. Transfer on the other metrics varies by
model.

\subsubsection{Additional Benchmarks and Answer Formats}
Table~\ref{tab:numerical} reports VSI-Bench question-type macro
accuracy~\cite{yang2025thinking} and SPBench mean relative accuracy
(MRA) on single-image and multi-view numerical questions. All three
metrics are reported separately from the five-metric average.

\begin{table}[!htbp]
\caption{\textbf{Additional external metrics (\%).} VSI macro averages
over question types. SI and MV MRA score single-image and multi-view
numerical answers on SPBench. Higher values are better in each column.}
\label{tab:numerical}
\centering
\footnotesize
\setlength{\tabcolsep}{2pt}
\begin{tabular*}{\columnwidth}{@{\extracolsep{\fill}}lrrr@{}}
\toprule
Model & VSI macro & SI MRA & MV MRA \\
\midrule
\multicolumn{4}{l}{\emph{Qwen3-VL-8B}} \\
Base~\cite{qwen3vl2025} & 43.12 & 56.11 & 65.97 \\
SpatialCast-GRPO & 43.33 & 56.50 & 65.43 \\
\midrule
\multicolumn{4}{l}{\emph{Qwen3-VL-4B}} \\
Base~\cite{qwen3vl2025} & 39.93 & 54.26 & 64.81 \\
Spatial-SSRL~\cite{spatialssrl2026} & 39.55 & 55.56 & 67.56 \\
SpatialBlock~\cite{spatialblock2026} & 37.21 & 45.29 & 46.12 \\
SpatialCast-GRPO & 39.45 & 52.34 & 64.15 \\
\midrule
\multicolumn{4}{l}{\emph{Qwen2.5-VL-3B}} \\
Base~\cite{qwen25vl2025} & 24.22 & 32.11 & 28.26 \\
SpatialBlock~\cite{spatialblock2026} & 25.52 & 14.66 & 26.47 \\
Spatial-MLLM~\cite{spatialmllm} & 44.30 & 52.78 & 63.91 \\
SpatialLadder~\cite{spatialladder2025} & 39.71 & 58.04 & 69.42 \\
Spatial-SSRL~\cite{spatialssrl2026} & 25.33 & 33.40 & 26.55 \\
SpatialCast-GRPO & 23.96 & 34.07 & 33.84 \\
\midrule
\multicolumn{4}{l}{\emph{InternVL3-2B}} \\
Base~\cite{internvl32025} & 17.84 & 28.99 & 10.00 \\
SpatialCast-GRPO & 18.92 & 27.43 & 7.67 \\
\bottomrule
\end{tabular*}

\end{table}

\FloatBarrier

\end{document}